%% file: main.tex
\documentclass[10pt,twocolumn,letterpaper]{article}

\usepackage[pagenumbers]{cvpr} 

\input{preamble}

\definecolor{cvprblue}{rgb}{0.21,0.49,0.74}
\usepackage[pagebackref,breaklinks,colorlinks,citecolor=cvprblue]{hyperref}

\def\paperID{*****} 
\def\confName{CVPR}
\def\confYear{2024}

\title{Generative AI in Vision: A Survey on Models, Metrics and Applications}

\author{Gaurav Raut\\
University of Maryland\\
{\tt\small gauraut14@gmail.com}
\and
Apoorv Singh\\
Carnegie Mellon Univeristy\\
{\tt\small apoorv93singh@gmail.com}
}

\begin{document}
\maketitle
\input{sec/0_abstract}
\input{sec/1_intro}
\input{sec/2_Gen_models_in_vision}
\input{sec/3_Metrics}

\input{sec/4_Applications}

\input{sec/5_Future_directions}
\input{sec/6_conclusion}
{
    \small
    \bibliographystyle{ieeenat_fullname}
    \bibliography{main}
}


\end{document}

%% file: preamble.tex
\usepackage[dvipsnames]{xcolor}
\usepackage{amsmath}
\usepackage{multirow}


%% file: sec/0_abstract.tex
\begin{abstract}
Generative AI models have revolutionized various fields by enabling the creation of realistic and diverse data samples. Among these models, diffusion models have emerged as a powerful approach for generating high-quality images, text, and audio. This survey paper provides a comprehensive overview of generative AI diffusion and legacy models, focusing on their underlying techniques, applications across different domains, and their challenges. We delve into the theoretical foundations of diffusion models, including concepts such as denoising diffusion probabilistic models (DDPM) and score-based generative modeling. Furthermore, we explore the diverse applications of these models in text-to-image, image inpainting, and image super-resolution, along with others, showcasing their potential in creative tasks and data augmentation. By synthesizing existing research and highlighting critical advancements in this field, this survey aims to provide researchers and practitioners with a comprehensive understanding of generative AI diffusion and legacy models and inspire future innovations in this exciting area of artificial intelligence.

\end{abstract}

%% file: sec/1_intro.tex
\section{Introduction}
\label{sec:intro}
\begin{figure}[!ht]
  \centering
  \includegraphics[width=0.47\textwidth]{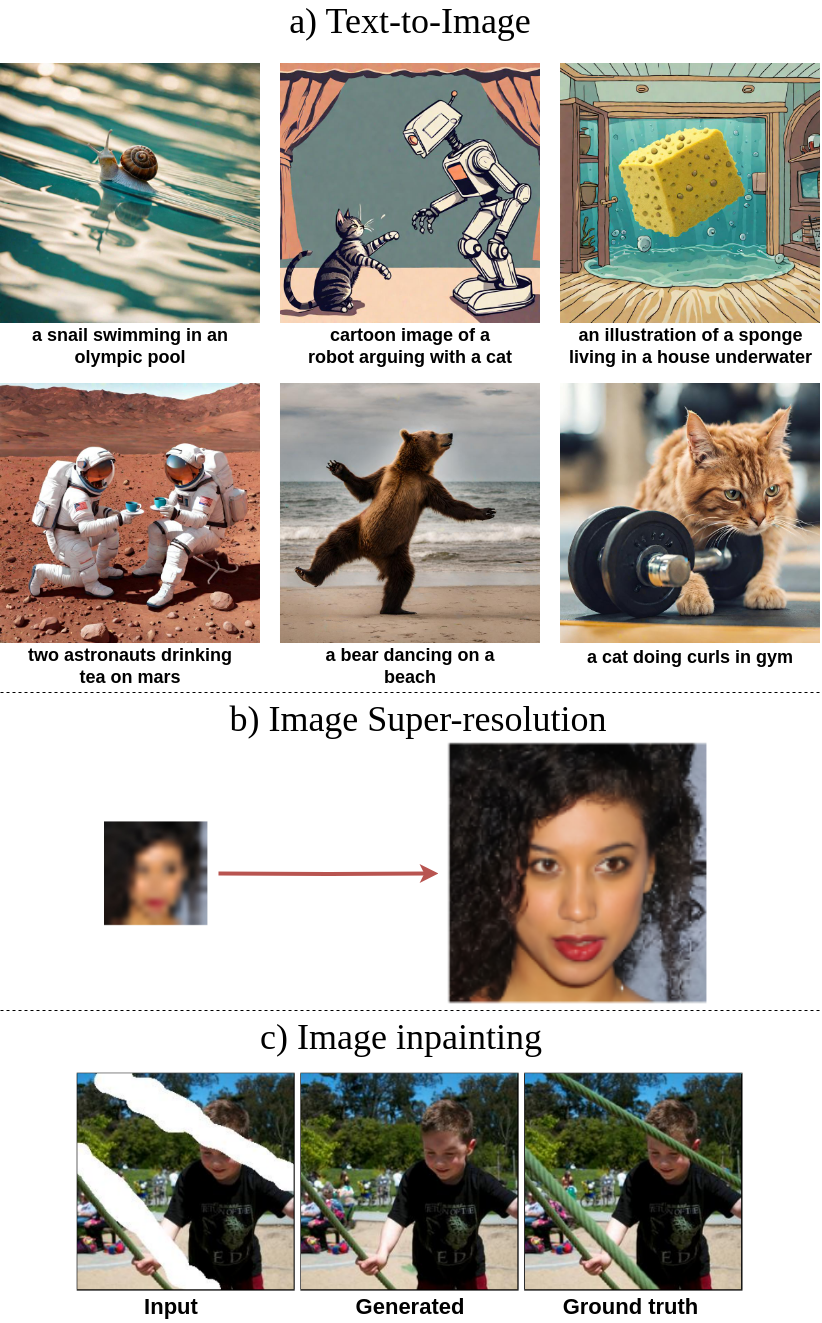}
  \caption{a) Images generated using stable diffusion\cite{rombach2022highresolution}; b) Image super-resolution results from SR3\cite{saharia2021image}; c) Image inpainting results from Palette\cite{saharia2022palette}}
  \label{fig:front}
\end{figure}

Generative models have long been at the forefront of artificial intelligence, enabling the creation of synthetic data samples with remarkable realism and diversity. Initially introduced as a method for denoising images, diffusion models have evolved to become a versatile framework for generating high-quality images, text, and audio data. Over the years, they have garnered significant attention from researchers and practitioners alike for their ability to capture complex data distributions and produce realistic samples. 

Generative models for computer vision started in 1950 through Hidden Markov models (HMMs) and Gaussian Mixture models (GMMs). These models used hand-designed features with limited complexities and diversity. With the advent of deep learning, Generative Adversarial Networks (GANs) and Variational Autoencoders (VAEs) enabled impressive image generation. However, in practice, GANs suffered several shortcomings in their architecture \cite{dhariwal2021diffusion}. The simultaneous training of generator and discriminator models was inherently unstable; sometimes, the generator “collapsed” and outputted lots of similar-seeming samples. Then came diffusion models, which were inspired by physics. Diffusion systems borrow from diffusion in non-equilibrium thermodynamics, where the process increases the entropy or randomness of the system over time. The recent innovation of diffusion models from OpenAI made them more practical in everyday applications. This paper dives into a systematic review of techniques and methodologies involved in SOTA diffusion models. 

The main contribution of this work can be summarized as follows: 
\begin{itemize}
    \item An overview of generative vision models to get readers up-to-speed with the theoretical prerequisites for going through the latest trends in diffusion models. 
    \item In-depth survey on the SOTA approaches for diffusion models, including 
    \item Highlight current research gaps and future research directions to provoke researchers to advance this generative vision modeling field further.
\end{itemize}

%% file: sec/2_Gen_models_in_vision.tex
\section{Generative Models in Vision}
\label{sec:models}
\begin{figure}[h]
  \centering
  \includegraphics[width=\linewidth]{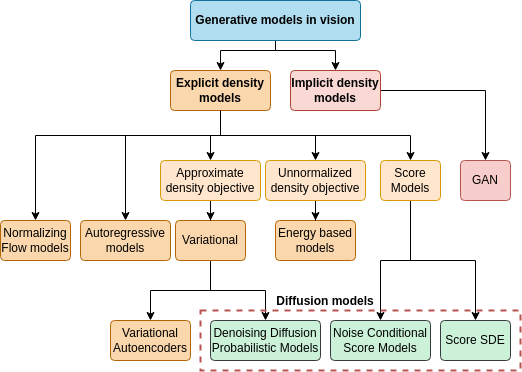}
  \caption{An extension of generative models classification based on\cite{goodfellow2017nips}}
  \label{fig:gen_mod_cat}
\end{figure}

Generative models in vision can be mainly classified into two categories \cref{fig:gen_mod_cat}: models that explicitly learn the probability density function by maximizing the likelihood or implicitly, where the model doesn't directly target to learn the density but does so using other strategies.\\
Normalizing flow and autoregressive models are dense probability estimators that model the exact likelihood of a distribution. Although, they are limited by the complexity of data distribution as converging the objective to achieve exact density representation of high dimensional complex data, such as images, can yield to computationally heavy and impractical models.\\
Variational autoencoders (VAE) alleviate the computation issue by allowing approximation of the intractable density distribution. This allows for a more efficient generative model with a trade-off of struggling against capturing complex data distributions.\\
Energy-based models offer a flexible modeling objective without any restrictions. They use an unnormalized representation of the probability distribution, making them excellent density estimators. However, the intractable objective makes them computationally inefficient for both training and sampling.\\
GANs\cite{goodfellow2014generative}, on the other hand, don't model the density objective directly. They rely on using an adversarial approach, which uses a minimax game between a discriminator and a generator to learn the density estimation explicitly. Although they have been largely successful on a wide set of applications, training them is difficult and suffer problems like vanishing gradient and mode collapse.\\
Diffusion models use variational or score-based approaches to model the probability density function. They work by perturbing data with continuous or discrete noise injection on either the data directly or a latent representation such as the latent diffusion model\cite{rombach2022highresolution} and learning the reverse denoising process.
\subsection{Diffusion models}
\subsubsection{Denoising Diffusion Probabilistic Models}
Denoising Diffusion Probabilistic Models, or DDPMs for short, are a class of diffusion models that are based on slowly introducing Gaussian noise to a training data sample $x_0$ over a large set of time steps ($1 \rightarrow T$), thereby obtaining a set of noise perturbed latent intermediates of the original sample ($x_1$, $x_2$, ..., $x_T$). This forward process is governed by the forward diffusion kernel (\textit{FDK}). At the end of $T$ time steps, we obtain the resulting $x_T$ sample, which can be approximated to represent isotropic Gaussian noise. From here, a deep neural network is tasked to learn the \textit{FDK}, and the parameters of this network are called the Reverse Diffusion Kernel (\textit{RDK}). The aim for \textit{RDK} is to predict the noise introduced by the \textit{FDK} at each time step starting from $x_T$ and slowly remove the noise to generate new samples that belong to the same probability distribution as our training dataset \cite{ho2020denoising, sohldickstein2015deep}.\\
\textbf{The forward process}: 
Let $q(\mathbf{x_0})$ represent the given sample's probability distribution before the noise perturbation. We define a Markov chain with our \textit{FDK} defined as a Gaussian distribution,
\begin{equation}
\label{forwardpdfddpm}
   q(\mathbf{x}_{1, \ldots, T}) := \prod_{i=1}^{T} q(\mathbf{x}_t \mid \mathbf{x}_{t-1})
\end{equation}
\begin{equation}
\label{fdk}
    q(\mathbf{x}_t \mid \mathbf{x}_{t-1}) := \mathcal{N}(\mathbf{x}_t;\sqrt{1-\beta_{t}} \mathbf{x}_{t-1}, \beta_{t}\mathrm{I})
\end{equation}
\begin{equation}
    \label{xforwardddpm}
    \mathbf{x}_{t} = \sqrt{1-\beta_{t}} \mathbf{x}_{t-1} + \sqrt{\beta_{t}} \epsilon 
\end{equation}
where, \cref{fdk} defines the \textit{FDK}; $\epsilon \sim \mathcal{N}(0,\mathbf{I})$. The term $\beta_{t}$ is a hyper-parameter in the diffusion process controlled by the variance scheduler. By applying this kernel to our $q(\mathbf{x}_0)$ repeatedly over $T-1$ time steps, we obtain $q(\mathbf{x}_{T})$ which approximates to an isotropic Gaussian distribution given that our covariance matrix in the \textit{FDK} is isotropic. Since the process is a Markov chain, we can seamlessly obtain any latent representation and probability distribution from $\mathbf{x}_0$ by simply substituting $\alpha_{t} := 1-\beta_{t}$ and $\overline{\alpha}_{t} := \prod_{s=1}^{t}\alpha_{s}$.
\begin{align}
    q({\mathbf{x}_{t} \mid \mathbf{x}_{0}}) := \mathcal{N}(\mathbf{x}_{t}; \sqrt{\overline{\alpha}_{t}}\mathbf{x}_{t-1}, (1-\overline{\alpha}_{t}\mathbf{I}) \\
    \mathbf{x}_{t} := \sqrt{\overline{\alpha}_{t}}\mathbf{x}_{0} + \sqrt{1-\overline{\alpha}_{t}}\epsilon
\end{align}
\textbf{The reverse process}: 
Starting from approximately an isotropic Gaussian distribution obtained at the time step $T$, the aim is to learn the \textit{RDK} $p_{\theta}(\mathbf{x}_{t-1} \mid \mathbf{x}_{t})$ which will predict the noise injected at each time step and will generate our original sample $q(\mathbf{x}_0)$ back within the finite $T$ time steps. The fact that the reverse process starts from a random isotropic Gaussian distribution serves as a cue that a trained network can produce new samples in the probability distribution of the training dataset while sampling.\\
The probability distribution in the reverse process to obtain our original sample can be illustrated as all the possible paths are taken from $p_{\theta}(\mathbf{x}_{T})$ to obtain $p_{\theta}(\mathbf{x}_{0}) \sim q(\mathbf{x}_{0})$ at each time step in $T \rightarrow 1$.
\begin{equation}
    \label{reversepdf}
        p_{\theta}(\mathbf{x}_{0}) := \int p_{\theta}(\mathbf{x}_{0 \ldots T} d\mathbf{x}_{0 \ldots T})
\end{equation}
This integral is intractable since it integrates over a complex high-dimensional space. To solve this, the authors \cite{ho2020denoising, sohldickstein2015deep} introduced a variational lower bound (or Evidence lower bound (\textbf{ELBO})) of the negative log-likelihood similar to VAEs\cite{kingma2022autoencoding} to minimize this. Using this, we obtain the variational lower bound loss, which is \cite{Croitoru_2023},

\begin{equation}
\begin{split}
   L_{VLB} = & \underbrace{-\log p_{\theta}(\mathbf{x}_{0} \mid \mathbf{x}_{1})}_{\mathbf{L}_{0}} + \underbrace{\mathbf{D}_{KL}(p(\mathbf{x}_{T} \mid \mathbf{x}_{0}) \| p_{\mathbf{x}_{T}})}_{\mathbf{L}_{T}} \\
   & + \underbrace{\sum_{t>1} \mathbf{D}_{KL}(p(\mathbf{x}_{t-1} \mid \mathbf{x}_{t}, \mathbf{x}_{0}) \| p_{\theta}(\mathbf{x}_{t-1} \mid \mathbf{x}_{t}))}_{\mathbf{L}_{t-1}},
\end{split}
\end{equation}
Here, we observe that the $\mathbf{L}_{T}$ doesn't depend on any learnable parameters from the network and hence can be omitted from the loss. The term $\mathbf{D}_{KL}$ is called the Kullback-Leibler divergence, a non-symmetric measure of the statistical distances between two probability distributions. This loss function trains the network and estimates the forward process posterior. We then define that the \textit{RDK} learned by the network is a Gaussian distribution given as $p_{\theta}(\mathbf{x}_{t-1} \mid \mathbf{x}_{t}) = \mathcal{N}(\mathbf{x}_{t-1}; \mu_{\theta}(\mathbf{x}_{t},t),\sigma_{\theta}(\mathbf{x}_{t},t))$ where the network, minimizing the variation lower bound loss, will learn to estimate the mean $\mu_{\theta}$ and the covariance $\sigma_{\theta}$ of the forward process posterior distribution. The authors further minimize the loss function in \cite{ho2020denoising} where they set the variance $\beta$ from \cref{xforwardddpm} as a constant. The authors then re-parameterize the mean $\mu_{\theta}$ in terms of noise, and we obtain the simplified loss function given as:
\begin{equation}
    L_{simple} := \mathbb{E}_{\mathbf{x}_{0},\epsilon} [\| \epsilon - \epsilon_{\theta}(\mathbf{x}_{t},t)\|^{2}],
\end{equation}
which is equivalent to the loss introduced in the score-based models\cite{song2020generative}.

\subsubsection{Noise Conditional Score Models}
Score-based models define the dataset's probability distribution as $q(\mathbf{x})$. The score function for the probability function can then be defined as the gradient of the log of this probability distribution, $\nabla_{\mathbf{x}} \log q(\mathbf{x})$ \cite{song2020generative}. The objective is to train a deep neural network, parameterized by $\theta$ to approximate over the score function of our dataset's probability function, $\mathbf{s}_{\theta}(\mathbf{x}) \sim \nabla_{\mathbf{x}} \log q(\mathbf{x})$, also known as score matching. Using the score function instead of the probability distribution allows the network to work with a tractable objective by eliminating the normalizing constant \cite{hyvarinen2005estimation}. The loss function is the Fischer divergence between the actual score and the learned score, which is derived to form,
\begin{equation}
    \label{loss_ncsm}
    \mathbb{E}_{q(\mathbf{x})}[\mathrm{tr}(\nabla_{\mathbf{x}}\mathbf{s}_{\theta}(\mathbf{x}))+\frac{1}{2}\|\mathbf{s}_{\theta}(\mathbf{x})\|^{2}]
\end{equation}
With a trained network, Langevin dynamics allows us to generate new samples using only the score function\cite{song2020generative}.
\begin{equation}
\label{sampling_langevin}
    \Tilde{\mathbf{x}}_{t} = \Tilde{\mathbf{x}}_{t-1} + \frac{\epsilon}{2}\nabla_{\mathbf{x}}\log q(\Tilde{\mathbf{x}}_{t-1}) + \sqrt{\epsilon}\mathcal{N}(0,\mathbf{I})
\end{equation}
Under the ideal conditions of $t \rightarrow \infty$ and $\epsilon \rightarrow 0$, we can generate exact samples coming from our dataset's distribution $q(\mathbf{x})$ \cite{welling2011bayesian}. The generation of new samples can then happen by simply substituting the learned score function $s_{\theta}(\mathbf{x})$ in \cref{sampling_langevin}.\\
However, the authors \cite{song2020generative} observed that this approach did not do well in practice because the scores generated were often inaccurate in lower-density regions. To address this, two solutions are employed to enhance the score matching using either the denoising score matching \cite{song2020generative} or sliced scored matching \cite{song2019sliced} where the loss in \cref{loss_ncsm} is either bypassed or approximated using random projections.\\
The idea of denoising score matching is to train the model by perturbing the dataset by inserting an isotropic Gaussian noise $\mathcal{N}(0,\sigma_{(1, \ldots, L)}\mathbf{I})$ and $\sigma_{1} < \ldots < \sigma_{L}$ where the prior data distribution is approximately equal to the noise perturbed distribution as the noise is inserted gradually over a large number of steps, $\nabla_{\mathbf{x}}\log q_{\sigma}(\mathbf{x}) \sim \nabla_{\mathbf{x}}\log q(\mathbf{x})$. This results in the modified loss function of,
\begin{equation}
    \label{dsmloss}
    \sum_{i=1}^{L}\lambda(i)\mathbb{E}_{q_{\sigma_{i}}(\mathbf{x})}[\|\nabla_{\mathbf{x}}\log q_{\sigma_{i}}(\mathbf{x}) - s_{\theta}(\mathbf{x},i)\|^{2}]
\end{equation}
where, $\lambda(i) \sim \sigma_{i}^{2}$ is a positive weighting function. For sampling, the Langevin dynamics are updated such that the trained network will produce samples similar to \cref{sampling_langevin} for each $i = L, \ldots, 1$, and the prior output will be used as the input for the next run. This process is called the annealed Langevin dynamics, producing a less noisy sample after every run from $i = L, \ldots, 1$.
\subsubsection{Stochastic Differential Equations Generative Models}
So far, we have seen that the diffusion models perturb data over a range of time steps or iterations $1 \rightarrow L$ \cite{ho2020denoising, song2020generative, song2020improved}. However, in \cite{song2021scorebased}, the authors generalized the previous approaches by defining a stochastic differential equation (SDE) to perturb data with noise in a continuum. The aim is to govern the diffusion process in both forward and reverse direction as a representation of an SDE.
\begin{equation}
    d\mathbf{x} = \mathbf{f}(\mathbf{x},t)dt + g(t)d\mathbf{w}
\end{equation}
where, $\mathbf{w}$ signifies the Brownian motion, the function $\mathbf{f}(\mathbf{x},t)$ is called as the drift coefficient and $g(t)$ is the diffusion coefficient \cite{song2021scorebased}. This is the SDE defining the forward process. Intuitively, it can be considered an SDE that forces a random sample to converge in areas of high probability densities. To sample, we need to solve this SDE backward in time. The reverse process is also a SDE\cite{anderson1982reverse, song2021scorebased} which is given as,
\begin{equation}
    d\mathbf{x} = [\mathbf{f}(\mathbf{x},t)-g(t)^{2}\nabla_{x}\log q_{t}(\mathbf{x})]dt + g(t)d\overline{\mathbf{w}}
\end{equation}
This SDE has a negative time step since the solution must be reversed ($t=T \rightarrow t=0$). To sample from this SDE, we need to learn the score as defined in \cite{song2020generative} as a function of time. Analogous to the noise conditional models, where the score function depends on the noise scales $\sigma_{L, \ldots, i, \ldots, 1}$, the score function here will depend on time $s_{\theta}(\mathbf{x}, t)$, formulating the loss function as,
\begin{equation}
    \mathbb{E}_{t}\mathbb{E}_{q_{t}(\mathbf{x})}[\lambda(t)\|\nabla_{\mathbf{x}}\log q_{t}(\mathbf{x}) - s_{\theta}(\mathbf{x},t)\|^{2}]
\end{equation}
After training our network $s_{\theta}(\mathbf{x}, t) \sim \nabla_{x}\log q_{t}(\mathbf{x})$, we can generate a new sample by solving the reverse SDE starting from pure noise $q_{T}(\mathbf{x})$. Many SDE solvers exist, the simplest of which is the Euler-Maruyama method. Similar to \cite{song2020generative}, we choose an infinitesimally small $\Delta(t)$ to solve the generalization of this SDE.
\begin{equation}
\begin{aligned}
    &\Delta(\mathbf{x}) = [\mathbf{f}(\mathbf{x},t)-g^{2}(t)s_{\theta}(\mathbf{x},t)]\Delta(t) + g(t)\sqrt{|\Delta t}|z_{t} \\
    &\mathbf{x} = \mathbf{x} + \Delta \mathbf{x} \\
    &t = t + \Delta t
\end{aligned}
\end{equation}
Following this, the authors improvise the sampling by introducing a predictor-corrector sampler where the idea is that for every step, the SDE solver will predict $\mathbf{x}(t+\Delta t)$ and then, the corrector will use this as an initial sample to refine the sample using $s_{\theta}(\mathbf{x}, t+\Delta t)$ by running it through the corrector network.
\subsection{Generative Adversarial Networks}
Generative Adversarial Networks (\textit{a.k.a.} GANs)\cite{goodfellow2014generative} are a class of generative models that implicitly learns the probability distribution $q(\mathbf{x})$ of the dataset using an adversarial approach where two networks, the discriminator $D$ and the Generator $G$ play a two-player min-max game. The discriminator's objective is to maximize the binary classification of distinguishing real and generated images, whereas the generator's objective is to fool the discriminator into misclassifying the generated images.
\begin{equation}
\label{vanGANloss}
\begin{aligned}
    &\min_{G} \max_{D} V(D,G) = \\
    &\mathbb{E}_{\mathbf{x}\sim q_{\mathbf{x}}}[\log D(\mathbf{x})] + \mathbb{E}_{\mathbf{z}\sim p_{\mathbf{z}}(\mathbf{z})}[\log(1-D(G(\mathbf{z})))],
\end{aligned}
\end{equation}
GANs are notoriously difficult to train \cite{goodfellow2014generative, salimans2016improved, radford2016unsupervised, arjovsky2017principled}. The objective is to find the Nash equilibrium (also referred to as a saddle point in other literature) of a two-player game ($D$ \& $G$) where both networks try to minimize their cost function simultaneously \cref{vanGANloss}. This results in a highly unstable training process as optimizing $D$ can lead to the deterioration of $G$ and vice-versa. Mode collapse is another problem where the generator's objective function converges to a specific data distribution instead of the whole training set, thus only generating images belonging to this small subset. Also, when the discriminator is trained to optimality, the gradients of $D$ approach zero. This causes the problem of vanishing gradients for the generator, where it has no guidance into which direction to follow for achieving optimality.\\
To prune these problems, various modifications on the vanilla GAN were introduced, which either suggested architectural optimizations or loss function optimizations \cite{Iglesias_2023}. As studied by \cite{Yi_2019,Creswell_2018}, the loss function optimizations were divided into optimizing the discriminator's $D$ or generator's $G$ loss objective. These include minimizing the f-divergences along with the Jensen-Shannon divergence \cite{nowozin2016fgan}, weight normalization for stabilizing $D$'s training \cite{miyato2018spectral}, WGAN and WGAN-GP\cite{arjovsky2017wasserstein, gulrajani2017improved} which changed the objective of $D$ from binary classification to a probability output by applying the Earth Mover (EM) or Wasserstein distance, EBGAN \cite{zhao2017energybased} introduces an energy-based formulation of $D$'s objective function where the architecture of $D$ is modified to be an auto-encoder, BEGAN \cite{berthelot2017began} which uses the same auto-encoder architecture for $D$ from EBGAN but modifies the objective to use the Wasserstein distance instead, SAGAN\cite{zhang2019selfattention} introduces self-attention modules on both $D$ and $G$ to enhance feature maps and uses spectral normalization\cite{miyato2018spectral}.\\

\subsection{Variational Autoencoders}
Variational Autoencoders (VAEs)\cite{Kingma_2019, kingma2022autoencoding} are generative models based on learning the latent space representation by projecting a prior $\mathbf{z}$ on the latent vector before generating a distribution. They have an encoder-decoder architecture similar to auto-encoders, while mathematically, they differ a lot \cite{doersch2021tutorial}. Instead of learning the latent vector as a discrete representation of the dataset, VAEs learn the probability distribution of this space. More intuitively, $q(\mathbf{x})$ is mapped to learn a multi-variate Gaussian distribution represented by the mean $\mu_{\mathbf{z}}$ and co-variance $\sigma_{\mathbf{z}}$ where $\mathbf{z}$ is in the latent space. While generating new samples, we want to start with the latent representation of $\mathbf{z}$ as an isotropic Gaussian distribution $\mathcal{N}(\mathbf{z}|\mu,\sigma*I)$. The regularization term and the reconstruction loss are introduced to achieve this. This regularization term is the KL divergence between the encoder's estimation of the latent variables and the standard Gaussian distribution. The loss function is then formulated as,
\begin{equation}
\begin{split}
    L(\theta,\phi;\mathbf{x}) = &\underbrace{-\mathbf{D}_{KL}(p_{\phi}(\mathbf{z}|\mathbf{x})||p_{\theta}(\mathbf{z}))}_{L_{KL}}+ \\&\underbrace{\mathbb{E}_{p_{\phi}(\mathbf{z}|\mathbf{x})}[\log p_{\theta}(\mathbf{x}|\mathbf{z})]}_{L_{reconstruction}}
\end{split}
\end{equation}
where $\phi$ and $\theta$ represent the encoder and decoder parameters, respectively. The only problem here is that $p_{\theta}(\mathbf{z})$ is intractable. To make the latent variable a learnable parameter, the authors \cite{kingma2022autoencoding} introduced a reparameterization trick to allow backpropagation. This is done by separating the stochastic part $\epsilon$ and reconstructing the latent vector as $\mathbf{z} = \mu + \sigma*\epsilon$. When the network has been trained to optimality, $p_{\theta}(\mathbf{x}|\mathbf{z}) \sim q(\mathbf{x})$.
\subsection{Autoregressive models}
Autoregressive models are generative models that use sequential data to calculate the likelihood of the next value in series\cite{pmlr-v15-larochelle11a, uria2016neural}. The joint distribution of a dataset can be given as,
\begin{equation}
    q(\mathbf{x}) = \prod_{i=1}^{N}q(\mathbf{x}_{i}|\mathbf{x}_{1},\ldots,\mathbf{x}_{n<i})
\end{equation}
In vision, this translates to generating images by sequencing the generation of each pixel given the prior pixels\cite{oord2016pixel, oord2016conditional, salimans2017pixelcnn}. More formally, the autoregressive network consists of either recurrent or convolutional layers that jointly learn the dataset's density distribution in a tractable manner and, during inference, will run $N = n^2$ times for an image of size $n*n$ to generate a sample. Images are not unlike audio\cite{oord2016wavenet} or text\cite{graves2014generating,vaswani2023attention} where the data is structured and sequenced. \cite{oord2016pixel} introduces a sequential approach for image synthesis by masking the pixels on the right and below and only considering the pixels above and on the left of the pixel we want to predict.
\subsection{Normalizing flow models}
Normalizing flow is a way of mapping a data's complex probability distribution $q(\mathbf{x})$ to a simple latent distribution $p(\mathbf{z})$ using a set of invertible, bijective and continuous functions $\mathbf{z} = f(\mathbf{x})$ such that both $f$ and $f^{-1}$ are differentiable\cite{rezende2016variational,dinh2015nice,dinh2017density}. When the function $f$ is a deep neural network, the model is called a normalizing flow model. Using the rule of change of variables, the probability density can be explicitly given as,
\begin{equation}
    p_{\mathbf{x}}(\mathbf{x},\theta) = p_{\mathbf{z}}(f_{\theta}(\mathbf{x})) \biggr\lvert\det \frac{\delta f_{\theta}(\mathbf{x})}{\delta \mathbf{x}}\biggr\rvert
\end{equation}
Since normalizing flow models estimate the exact likelihood of the distribution, the training is done by minimizing the negative log-likelihood given as,
\begin{equation}
    L_{NF} = -\log p_{\mathbf{x}}(\mathbf{x},\theta) = -\log p_{\mathbf{z}}(f_{\theta}(\mathbf{x})) - \log \biggr\lvert\det \frac{\delta f_{\theta}(\mathbf{x})}{\delta \mathbf{x}}\biggr\rvert
\end{equation}
When the model is trained, the latent representation $\mathbf{z}$, often chosen as a multivariate Gaussian distribution\cite{Kobyzev_2021,papamakarios2021normalizing}, can generate a sample from the dataset's probability distribution by simply applying the inverse function $f_{\theta}^{-1}$ to it. Although flow models are based on modeling the exact data distribution, they are often computationally expensive since they depend on the calculation of jacobians, have scalability and expressiveness issues on large and complex data distributions, and because of the invertibility constraint in calculations, require the input and output dimensions to be the same\cite{Bond_Taylor_2022}.
\subsection{Energy Based models}
In simplicity, Energy-based models (EBMs) \cite{yann,jiquan,song2021train} are generative models that assign an energy function $E_{\theta}(\mathbf{x})$ to a dataset's probability distribution $q(\mathbf{x})$ and minimize the energy function for samples from the dataset while assigning high energy to samples that don't belong to it.
\begin{equation}
\label{ebmll}
    p_{\theta}(\mathbf{x}) = \frac{e^{-E_{\theta}(\mathbf{x})}}{Z_{\theta}}
\end{equation}
EBMs are incredibly flexible in the domain of the type of data\cite{duebmimplicit, song2021train, jiquan} and, in vision, are an excellent choice for tasks of anomaly detection as an optimally trained model can distinguish between the anomalies and ideal sample\cite{zhai2016deep}.\\
Because the energy function is unnormalized, $Z_{\theta} = \int_{\mathbf{x}}e^{-E_{\theta}(\mathbf{x})}d\mathbf{x}$ is used in \cref{ebmll} for normalizing the likelihood which is often intractable. This makes the training and sampling of EBMs difficult, and one has to rely on computationally heavy methods such as MCMC, contrastive divergence (CD), and score matching, which make them an impractical choice for fast inference use cases.

%% file: sec/3_Metrics.tex
\section{Evaluation metrics for vision generative models}
\label{metrics}
Evaluating generative models in vision is an active research topic where different tasks involve specialized metrics such as DrawBench\cite{saharia2021image}, PartiPrompts\cite{yu2022scaling}, CLIPScore\cite{hessel2022clipscore} for text-to-image tasks or PSNR for image reconstruction tasks. Here, we mainly introduce metrics that measure image fidelity and model diversity.
\subsection{Inception Score (IS)}
Inception Score (or IS) was first introduced to assess the quality of the images generated by GAN as an automated alternative process against human annotators\cite{salimans2016improved}. Using a feature extracting network (often the Inception model\cite{szegedy2014going}), which is trained on the same dataset as the generative model, the score measures two components of the generated samples: entropy of a single sample over the class labels and secondly, entropy of class distribution over a large number of samples (suggested close to 50K samples) to measure the diversity. For a well-trained model, the entropy of a class over a single sample should be low, and the entropy of class distribution over all generated samples should be high. This indicates that the network can generate both meaningful and diverse sets of images. The score is calculated as follows:
\begin{equation}
    \mathbf{IS} = \exp(\mathbb{E}_{\mathbf{x}}[\mathbf{D}_{KL}(p_{\theta}(\mathbf{y}|\mathbf{x})\|p_{\theta}(\mathbf{y})))]
\end{equation}
where $\mathbf{D}_{KL}$ is the KL divergence, $p_{\theta}(\mathbf{y}|\mathbf{x})$ is the predicted class probability and $p_{\theta}(\mathbf{y})$ is the probability of classes over all generated images. This implies that the higher the IS score is, the better the model's generative capabilities.
\subsection{Fr\'echet Inception Distance}
The drawback of the Inception score is that it only considers the generated samples for evaluation and disregards comparing them with the actual dataset. Also, the IS will only compare the class probabilities as opposed to the image feature distribution, which causes it to miss the more relevant image features and requires a labeled dataset. The Fr\'echet Inception Distance (FID)\cite{DOWSON1982450,heusel2018gans} allows us to overcome this by comparing the extracted features from a certain layer of the feature extractor. More specifically, it assumes that the extracted features belong to a Gaussian distribution with a certain mean $\mu$ and co-variance $\Sigma$ and calculates the Fr\'echet distance (measures the similarity between two probability distributions) between samples generated by the model ($\mu_{g}$,$\Sigma_{g}$) and the samples from the training ($\mu_{t}$,$\Sigma_{t}$) dataset($\sim$ 50K samples). It is given as,
\begin{equation}
    \mathbf{D}_{FID} = \|\mu_{t} - \mu_{g}\|^{2} + \mathrm{tr}(\Sigma_{t} + \Sigma_{g} - 2\sqrt{\Sigma_{t}\Sigma_{g}})
\end{equation}
Since FID is a similarity distance, the lower the FID is, the better.
For zero-shot applications \cite{glide,dalle2,parti}, a modified version called the zero-shot FID is used where $\mu_{t},\Sigma_{t}$ signify the target distribution based on textual cues as opposed to the training dataset.
Kernel Inception Distance (KID) \cite{kid} is another flavor of the FID, where the distance is measured between a polynomial representation of the inception layer's distribution.
\subsection{Precision and Recall}
Precision and recall \cite{sajjadi2018assessing,kynkäänniemi2019improved} are two metrics that follow the same motivation as IS and FID of measuring the quality and diversity of the generated samples while overcoming issues of mode dropping. It provides a two-dimensional score where precision measures the quality of images produced while recall measures the diversity coverage of the generative model. 

%% file: sec/4_Applications.tex
\section{Applications in Vision}
\label{applications}

Generative models find applications in various tasks such as image denoising, inpainting, super-resolution, text-to-video synthesis, image-to-image translation, image search, and reverse image search. These applications can be divided into two broad categories:

\subsection{Unconditional generation}
As the name suggests, unconditional generative models are trained to learn a target distribution and synthesize new samples without getting conditioned by any other input. All models described in \cref{sec:models} can be considered a base unconditioned model whose only focus is on learning the target distribution\cite{yang2023diffusion,Croitoru_2023}. Unconditional image generation models usually start with a seed that generates a random noise vector. The model will then use this vector to create output images that resemble training data distribution. 
\subsection{Conditional generation}
On the contrary, conditional diffusion models take a prompt and some random initial noise and iteratively remove the noise to construct an image. The prompt guides the denoising process, and once the denoising process ends after a predetermined number of time steps, the image representation is decoded into an image. There are several forms of conditional models:

\begin{figure}
    \centering
    \includegraphics[width=\linewidth]{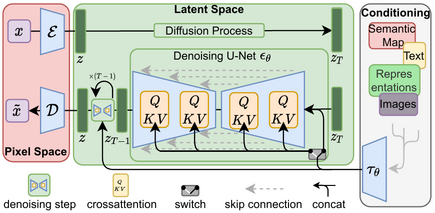}
    \caption{An example of a conditional generative model: Latent diffusion model\cite{rombach2022highresolution}}
    \label{fig:latentdiff}
\end{figure}
\subsubsection{Text-to-Image Generation}
Text-to-image has been the most naturally prominent use case for generative models. Providing conditioned textual information for image generation has improved the model's generative capabilities\cite{radford2021learning,zhang2023texttoimage}. Over the years, architectures reigning the task have used recurrent layers \cite{mansimov2016generating}, GANs \cite{reed2016generative, zhang2017stackgan, li2019controllable, sauer2023stylegant, kang2023scaling}, autoregressive models \cite{ramesh2021zeroshot, ding2021cogview, yu2022scaling} and the diffusion-based models \cite{gu2022vector,ramesh2022hierarchical,saharia2022photorealistic,rombach2022highresolution,chen2022reimagen}. Below we present an overview of their functioning and also present their metrics in \cref{tab:comparison}.

\begin{table}[h]
  \small
  \centering
  \caption{Comparison of different text-to-image models}
  \resizebox{\columnwidth}{!}{
  \begin{tabular}{|p{2cm}|p{2cm}|c|c|c|c|}
    \hline
    \textbf{Models} & \textbf{Architecture} & \multicolumn{4}{c|}{\textbf{Best Reported}}\\
    \cline{3-6}
     &  & \textbf{Zero-shot FID$\downarrow$} & \textbf{IS$\uparrow$} & \textbf{Params} & \textbf{Dataset}\\
    \hline
    StackGAN\cite{zhang2017stackgan} & GAN  & - & 8.45 & - & COCO\\
    GigaGAN\cite{kang2023scaling} & GAN & 9.09 & - & 1.0B & COCO\\
    DALL-E\cite{ramesh2021zeroshot} & Autoregressive & ~27.50 & ~18 & 12B & COCO\\
    GLIDE\cite{nichol2022glide} & Diffusion & 12.24 & ~23.7 & 5B & COCO\\
    Stable Diffusion\cite{rombach2022highresolution} & Latent Diffusion & 12.63 & 30.29 & 1.45B & COCO\\
    DALL-E2\cite{ramesh2022hierarchical} & Diffusion & 10.39 & - & 5.5B & COCO\\
    Imagen\cite{saharia2022photorealistic} & Diffusion & 7.27 & - & 3B & COCO\\
    Parti-20B\cite{parti} & Autoregressive & \textbf{3.22} & - & 20B & COCO\\
    Re-Imagen\cite{chen2022reimagen} & Diffusion & 6.88 & - & 3.6B & COCO\\
    \hline
  \end{tabular}
  }
  \label{tab:comparison}
\end{table}

\textbf{\textit{Text conditioned GANs}}: Followed by cGANs \cite{mirza2014conditional}, \cite{reed2016generative} introduced embedding textual information to achieve text-to-image generation. The model is jointly trained using images and text captions. During sampling, the text prompts are converted to text encoding using an encoder and compressed using a 128-dimensional fully connected layer. This compressed encoding is concatenated with the latent vector $\mathbf{z}$ and passed to the generator to generate images. StackGAN \cite{zhang2017stackgan} introduced a series of two GAN networks where the text encodings from the encoder are mapped to a Gaussian distribution with random noise and are then concatenated with the latent vector of the stage-I GAN. The stage-I GAN generates a low-dimensional image, which is further embedded in the latent vector of the stage-II GAN to generate a high-resolution image. The image compression is done using the trained discriminator from stage-I. \\
\textbf{\textit{Text conditioned autoregressive models}}: DALL-E \cite{ramesh2021zeroshot} is a two-stage autoregressive transformer model where the first stage has a discrete VAE to tokenize images in a 32$\times$32 grid and the second stage concatenates the text encodings from a BP-Encoder to create text tokens. These concatenated image-text tokens are jointly trained to maximize the ELBO \cite{kingma2022autoencoding} to obtain an image-text distribution. Similarly, CogView \cite{ding2021cogview} employs a VQ-VAE\cite{oord2018neural} for image tokenization and SentencePiece\cite{kudo2018sentencepiece} for generating text tokens and Parti\cite{yu2022scaling} uses the ViT-VQGAN\cite{yu2022vectorquantized} as the image tokenizer and a pre-trained BERT\cite{devlin2019bert} for text encodings with the best fine-tuned model achieving SOTA FID score of 3.22 on MS-COCO dataset.\\
\textbf{\textit{Text conditioned Diffusion models}}: GLIDE\cite{nichol2022glide} uses an ADM model \cite{dhariwal2021diffusion} and a transformer-based text encoder to generate prompts in place of class labels for image synthesis. DALL-E2 (\textit{a.k.a.} unCLIP)\cite{ramesh2022hierarchical} is a two-stage model where the first stage uses the CLIP\cite{radford2021learning} model to generate image embedding from text captions and the second stage uses these image embedding as a prior to generating samples via a diffusion decoder. The authors also experimented with using an autoregressive decoder instead of the diffusion decoder, but the latter yielded better results. Stable Diffusion\cite{rombach2022highresolution} is a latent diffusion model (LDM). Instead of directly dealing with the complex pixel representation, LDMs apply the DDPM model's forward kernel on the latent representation generated by the encoder \cref{fig:latentdiff}. A series of trained denoising U-Net are applied to denoise this corrupted latent space. The text prompts are embedded in the denoising steps using cross-attention layers. The retrieved latent space, after denoising, is passed through a decoder to generate the sample. Because LDMs use diffusion on the latent representation, training, and sampling have proven to be computationally inexpensive as compared to other models. Imagen\cite{saharia2022photorealistic} encodes its textual prompts using a T5-XXL LLM model similar to CLIP. It then generates low-resolution images (64$\times$64) using a series of denoising U-Net and then up-samples these images using a series of two super-resolution U-Net diffusion models to generate images of size 256$\times$256 and 1024$\times$1024 respectively. Re-Imagen\cite{chen2022reimagen} focuses on retrieving k-nearest neighboring images from a dataset based on the text prompts provided and uses these images as a reference to generate new samples. DALL-E3\cite{betker2023improving} attends to improving image quality by re-captioning text prompts into a more descriptive prompt, which has proven to generate higher-quality images.
\subsubsection{Image super resolution}
ViT-based models have been shown to achieve SOTA results for the task of image super-resolution\cite{chen2023activating,zhang2023swinfir,chen2023dual,chen2023recursive,zhang2024transcending}. Even so, the generalization capabilities of generative models can soon catch up the leaderboard\cite{Gao_2023_CVPR,10353979}. SRGAN\cite{ledig2017photorealistic} first introduced a generative framework for this task using an adversarial objective. ESRGAN\cite{wang2018esrgan} and RFB-ESRGAN\cite{shang2020perceptual} further improvise the SRGAN implementation by employing architectural modifications such as relativistic discriminator, dense residual block, and upgrading the perceptual loss. GLEAN \cite{Chan_2021_CVPR} introduced a novel encoder-bank-decoder approach where the encoder's latent vectors and multi-layer convolutional features are passed to a StyleGAN\cite{karras2019stylebased} based latent bank. This generative bank combines features from the encoder at various scales and generates new latent feature representations passed to the decoder to generate a super-resolution image. SR3\cite{saharia2021image} and SRDiff\cite{LI202247} were the first diffusion (DDPM) based SR models. SR3 uses a conditional DDPM U-Net architecture with some adaptations in the residual layers and conditions the denoising process using an LR image directly at each iteration. SRDiff uses an encoder-generated embedding of the LR image and conditions the denoising step by concatenating the embeddings at each iteration. IDM\cite{Gao_2023_CVPR} uses an implicit neural representation in addition to the conditioned DDPM to achieve a continuous restoration over multiple resolutions using the current iteration's features. EDiffSR\cite{10353979} uses the SDE diffusion process where isotropic noise is conditioned with the LR image during sampling to generate the high-resolution image.
\subsubsection{Image anomaly detection}
AnoGAN\cite{schlegl2017unsupervised} and f-AnoGAN\cite{SCHLEGL201930} are unsupervised adversarial anomaly detection networks that were trained on healthy data and using the proposed anomaly score along with the residual score; the model predicts anomalies on unseen data depending on the variation in the learned latent space. DifferNet\cite{rudolph2020differnet} uses the normalizing flow model to map the density of healthy image features extracted from a feature extraction network. By training on healthy data, anomalies will have a lower likelihood and will be out of distribution in the density space. FastFlow\cite{yu2021fastflow} uses a similar approach but extends the normalizing flow into a 2D space, which allows to directly output location results of anomalies. CFLOW-AD\cite{Gudovskiy_2022_WACV} uses an encoder feature extractor where features from every scale are pooled to form multi-scale feature vectors, which are passed to the specific normalizing flow decoder along with the positional encoding for localization of anomalies. The outputs from each decoder are aggregated to generate an anomaly map. AnoDDPM\cite{9857019} approaches the problem using a DDPM model with simplex noise for data perturbation. The diffusion model is trained to generate healthy images. During inference, the simplex noise perturbs the test sample for a certain number of pre-set steps, and the denoising diffusion model then generates an anomaly-free image using the perturbed sample as the prior. Comparing the generated and input samples using reconstruction error, an anomaly segmentation map is generated.
\subsubsection{Image inpainting}
Image inpainting tasks include restoration, textural synthesis, and mask filling\cite{10.1145/344779.344972}. Context encoders \cite{pathak2016context} applied an adversarial loss along with the reconstruction loss to achieve sharp and coherent mask filling. \cite{Iizuka} advances the simple discriminator by introducing a mixture of local and global context discriminators. The local discriminator has the filled mask as input, and the global discriminator takes the whole image as input to collectively create the discriminator's objective. Following it, \cite{Yu_2018_CVPR} uses a two-step generative approach where the first generator (trained on reconstruction loss) generates a coarse prediction and the second generator with a WGAN-GP\cite{gulrajani2017improved} based objective is the refinement generator trained on local and global adversarial loss along with the reconstruction loss. StructurFlow\cite{ren2019structureflow} bifurcates the GAN generation in two steps: the structure generator and the texture generator. The structure generator creates a smoothened edge-preserved image, and the texture generator fills the texture in the smooth reconstructed image. The texture generator uses an additional input of appearance flow in the latent space, which predicts the texture of the masked regions based on the texture from source regions. CoModGAN\cite{zhao2021large} further generalizes the inpainting task with both input-masked image and stochastic noise-conditioned latent vector input to the generator, which enabled them for large region image inpainting. Pallete\cite{saharia2022palette} is a DDPM\cite{ho2020denoising} based image-to-image translation model capable of image inpainting. It fills the masked region with standard Gaussian noise and performs the denoising training only on the masked region. RePaint\cite{Lugmayr_2022_CVPR} salvages the pre-trained unconditional DDPM model instead of training a model for the inpainting task. Since DDPM follows a Markov chain for data perturbation, the masked input image's noise-perturbed data is known for every iteration in reverse. Using this knowledge, the denoising process is conditioned to add the noise-perturbed image at each reverse step to predict the masked region. Given the stochastic nature, this process can generate multiple candidates for the inpainting task.
\subsubsection{Other tasks}
In addition to the above tasks, these models can be used for various other generative tasks like image-to-image translation\cite{isola2018imagetoimage,tumanyan2023plug},  image colorization\cite{saharia2022palette}, video generation\cite{ho2022imagen}, point cloud generation\cite{luo2021diffusion}, restoration, etc.

%% file: sec/5_Future_directions.tex
\section{Future directions}
Some of the unsolved but highly sought-after directions that researchers can take are: 
\begin{itemize}
    \item Exploring Time-Series Forecasting Applications: Future research could delve deeper into leveraging diffusion models for improved forecasting accuracy and efficiency.
    \item Physics-Inspired Generative Models: Future research could focus on advancing physics-inspired generative models to achieve unprecedented speed and quality in content creation.
    \item Ethical Considerations: Future research could involve addressing issues related to bias, fairness, and the societal impact of generative diffusion models.
\end{itemize}

%% file: sec/6_conclusion.tex
\section{Conclusion}
This survey paper has provided a detailed examination of generative AI diffusion models, shedding light on their techniques, applications, and challenges. Despite their promising capabilities, generative AI diffusion models still face significant challenges such as training stability, scalability issues, and interpretability concerns. Addressing these challenges will be crucial for advancing the field and unlocking the full potential of diffusion models in generating realistic and diverse data samples. By synthesizing current research findings and identifying key areas for future research, this survey aims to guide researchers and practitioners toward further advancements in generative AI diffusion models, paving the way for innovative applications and breakthroughs in artificial intelligence.

%% file: main.bbl
\begin{thebibliography}{114}
\providecommand{\natexlab}[1]{#1}
\providecommand{\url}[1]{\texttt{#1}}
\expandafter\ifx\csname urlstyle\endcsname\relax
  \providecommand{\doi}[1]{doi: #1}\else
  \providecommand{\doi}{doi: \begingroup \urlstyle{rm}\Url}\fi

\bibitem[Anderson(1982)]{anderson1982reverse}
Brian~DO Anderson.
\newblock Reverse-time diffusion equation models.
\newblock \emph{Stochastic Processes and their Applications}, 12\penalty0 (3):\penalty0 313--326, 1982.

\bibitem[Arjovsky and Bottou(2017)]{arjovsky2017principled}
Martin Arjovsky and Léon Bottou.
\newblock Towards principled methods for training generative adversarial networks, 2017.

\bibitem[Arjovsky et~al.(2017)Arjovsky, Chintala, and Bottou]{arjovsky2017wasserstein}
Martin Arjovsky, Soumith Chintala, and Léon Bottou.
\newblock Wasserstein gan, 2017.

\bibitem[Bertalmio et~al.(2000)Bertalmio, Sapiro, Caselles, and Ballester]{10.1145/344779.344972}
Marcelo Bertalmio, Guillermo Sapiro, Vincent Caselles, and Coloma Ballester.
\newblock Image inpainting.
\newblock In \emph{Proceedings of the 27th Annual Conference on Computer Graphics and Interactive Techniques}, page 417–424, USA, 2000. ACM Press/Addison-Wesley Publishing Co.

\bibitem[Berthelot et~al.(2017)Berthelot, Schumm, and Metz]{berthelot2017began}
David Berthelot, Thomas Schumm, and Luke Metz.
\newblock Began: Boundary equilibrium generative adversarial networks, 2017.

\bibitem[Betker et~al.(2023)Betker, Goh, Jing, Brooks, Wang, Li, Ouyang, Zhuang, Lee, Guo, et~al.]{betker2023improving}
James Betker, Gabriel Goh, Li Jing, Tim Brooks, Jianfeng Wang, Linjie Li, Long Ouyang, Juntang Zhuang, Joyce Lee, Yufei Guo, et~al.
\newblock Improving image generation with better captions.
\newblock \emph{Computer Science. https://cdn. openai. com/papers/dall-e-3. pdf}, 2\penalty0 (3):\penalty0 8, 2023.

\bibitem[Bińkowski et~al.(2021)Bińkowski, Sutherland, Arbel, and Gretton]{kid}
Mikołaj Bińkowski, Danica~J. Sutherland, Michael Arbel, and Arthur Gretton.
\newblock Demystifying mmd gans, 2021.

\bibitem[Bond-Taylor et~al.(2022)Bond-Taylor, Leach, Long, and Willcocks]{Bond_Taylor_2022}
Sam Bond-Taylor, Adam Leach, Yang Long, and Chris~G. Willcocks.
\newblock Deep generative modelling: A comparative review of vaes, gans, normalizing flows, energy-based and autoregressive models.
\newblock \emph{IEEE Transactions on Pattern Analysis and Machine Intelligence}, 44\penalty0 (11):\penalty0 7327–7347, 2022.

\bibitem[Chan et~al.(2021)Chan, Wang, Xu, Gu, and Loy]{Chan_2021_CVPR}
Kelvin~C.K. Chan, Xintao Wang, Xiangyu Xu, Jinwei Gu, and Chen~Change Loy.
\newblock Glean: Generative latent bank for large-factor image super-resolution.
\newblock In \emph{Proceedings of the IEEE/CVF Conference on Computer Vision and Pattern Recognition (CVPR)}, pages 14245--14254, 2021.

\bibitem[Chen et~al.(2022)Chen, Hu, Saharia, and Cohen]{chen2022reimagen}
Wenhu Chen, Hexiang Hu, Chitwan Saharia, and William~W. Cohen.
\newblock Re-imagen: Retrieval-augmented text-to-image generator, 2022.

\bibitem[Chen et~al.(2023{\natexlab{a}})Chen, Wang, Zhou, Qiao, and Dong]{chen2023activating}
Xiangyu Chen, Xintao Wang, Jiantao Zhou, Yu Qiao, and Chao Dong.
\newblock Activating more pixels in image super-resolution transformer, 2023{\natexlab{a}}.

\bibitem[Chen et~al.(2023{\natexlab{b}})Chen, Zhang, Gu, Kong, and Yang]{chen2023recursive}
Zheng Chen, Yulun Zhang, Jinjin Gu, Linghe Kong, and Xiaokang Yang.
\newblock Recursive generalization transformer for image super-resolution, 2023{\natexlab{b}}.

\bibitem[Chen et~al.(2023{\natexlab{c}})Chen, Zhang, Gu, Kong, Yang, and Yu]{chen2023dual}
Zheng Chen, Yulun Zhang, Jinjin Gu, Linghe Kong, Xiaokang Yang, and Fisher Yu.
\newblock Dual aggregation transformer for image super-resolution, 2023{\natexlab{c}}.

\bibitem[Creswell et~al.(2018)Creswell, White, Dumoulin, Arulkumaran, Sengupta, and Bharath]{Creswell_2018}
Antonia Creswell, Tom White, Vincent Dumoulin, Kai Arulkumaran, Biswa Sengupta, and Anil~A. Bharath.
\newblock Generative adversarial networks: An overview.
\newblock \emph{IEEE Signal Processing Magazine}, 35\penalty0 (1):\penalty0 53–65, 2018.

\bibitem[Croitoru et~al.(2023)Croitoru, Hondru, Ionescu, and Shah]{Croitoru_2023}
Florinel-Alin Croitoru, Vlad Hondru, Radu~Tudor Ionescu, and Mubarak Shah.
\newblock Diffusion models in vision: A survey.
\newblock \emph{IEEE Transactions on Pattern Analysis and Machine Intelligence}, 45\penalty0 (9):\penalty0 10850–10869, 2023.

\bibitem[Devlin et~al.(2019)Devlin, Chang, Lee, and Toutanova]{devlin2019bert}
Jacob Devlin, Ming-Wei Chang, Kenton Lee, and Kristina Toutanova.
\newblock Bert: Pre-training of deep bidirectional transformers for language understanding, 2019.

\bibitem[Dhariwal and Nichol(2021)]{dhariwal2021diffusion}
Prafulla Dhariwal and Alex Nichol.
\newblock Diffusion models beat gans on image synthesis, 2021.

\bibitem[Ding et~al.(2021)Ding, Yang, Hong, Zheng, Zhou, Yin, Lin, Zou, Shao, Yang, and Tang]{ding2021cogview}
Ming Ding, Zhuoyi Yang, Wenyi Hong, Wendi Zheng, Chang Zhou, Da Yin, Junyang Lin, Xu Zou, Zhou Shao, Hongxia Yang, and Jie Tang.
\newblock Cogview: Mastering text-to-image generation via transformers, 2021.

\bibitem[Dinh et~al.(2015)Dinh, Krueger, and Bengio]{dinh2015nice}
Laurent Dinh, David Krueger, and Yoshua Bengio.
\newblock Nice: Non-linear independent components estimation, 2015.

\bibitem[Dinh et~al.(2017)Dinh, Sohl-Dickstein, and Bengio]{dinh2017density}
Laurent Dinh, Jascha Sohl-Dickstein, and Samy Bengio.
\newblock Density estimation using real nvp, 2017.

\bibitem[Doersch(2021)]{doersch2021tutorial}
Carl Doersch.
\newblock Tutorial on variational autoencoders, 2021.

\bibitem[Dowson and Landau(1982)]{DOWSON1982450}
D.C Dowson and B.V Landau.
\newblock The fréchet distance between multivariate normal distributions.
\newblock \emph{Journal of Multivariate Analysis}, 12\penalty0 (3):\penalty0 450--455, 1982.

\bibitem[Du and Mordatch(2019)]{duebmimplicit}
Yilun Du and Igor Mordatch.
\newblock Implicit generation and modeling with energy based models.
\newblock In \emph{Advances in Neural Information Processing Systems}. Curran Associates, Inc., 2019.

\bibitem[Gao et~al.(2023)Gao, Liu, Zeng, Xu, Li, Luo, Liu, Zhen, and Zhang]{Gao_2023_CVPR}
Sicheng Gao, Xuhui Liu, Bohan Zeng, Sheng Xu, Yanjing Li, Xiaoyan Luo, Jianzhuang Liu, Xiantong Zhen, and Baochang Zhang.
\newblock Implicit diffusion models for continuous super-resolution.
\newblock In \emph{Proceedings of the IEEE/CVF Conference on Computer Vision and Pattern Recognition (CVPR)}, pages 10021--10030, 2023.

\bibitem[Goodfellow(2017)]{goodfellow2017nips}
Ian Goodfellow.
\newblock Nips 2016 tutorial: Generative adversarial networks, 2017.

\bibitem[Goodfellow et~al.(2014)Goodfellow, Pouget-Abadie, Mirza, Xu, Warde-Farley, Ozair, Courville, and Bengio]{goodfellow2014generative}
Ian~J. Goodfellow, Jean Pouget-Abadie, Mehdi Mirza, Bing Xu, David Warde-Farley, Sherjil Ozair, Aaron Courville, and Yoshua Bengio.
\newblock Generative adversarial networks, 2014.

\bibitem[Graves(2014)]{graves2014generating}
Alex Graves.
\newblock Generating sequences with recurrent neural networks, 2014.

\bibitem[Gu et~al.(2022)Gu, Chen, Bao, Wen, Zhang, Chen, Yuan, and Guo]{gu2022vector}
Shuyang Gu, Dong Chen, Jianmin Bao, Fang Wen, Bo Zhang, Dongdong Chen, Lu Yuan, and Baining Guo.
\newblock Vector quantized diffusion model for text-to-image synthesis, 2022.

\bibitem[Gudovskiy et~al.(2022)Gudovskiy, Ishizaka, and Kozuka]{Gudovskiy_2022_WACV}
Denis Gudovskiy, Shun Ishizaka, and Kazuki Kozuka.
\newblock Cflow-ad: Real-time unsupervised anomaly detection with localization via conditional normalizing flows.
\newblock In \emph{Proceedings of the IEEE/CVF Winter Conference on Applications of Computer Vision (WACV)}, pages 98--107, 2022.

\bibitem[Gulrajani et~al.(2017)Gulrajani, Ahmed, Arjovsky, Dumoulin, and Courville]{gulrajani2017improved}
Ishaan Gulrajani, Faruk Ahmed, Martin Arjovsky, Vincent Dumoulin, and Aaron Courville.
\newblock Improved training of wasserstein gans, 2017.

\bibitem[Hessel et~al.(2022)Hessel, Holtzman, Forbes, Bras, and Choi]{hessel2022clipscore}
Jack Hessel, Ari Holtzman, Maxwell Forbes, Ronan~Le Bras, and Yejin Choi.
\newblock Clipscore: A reference-free evaluation metric for image captioning, 2022.

\bibitem[Heusel et~al.(2018)Heusel, Ramsauer, Unterthiner, Nessler, and Hochreiter]{heusel2018gans}
Martin Heusel, Hubert Ramsauer, Thomas Unterthiner, Bernhard Nessler, and Sepp Hochreiter.
\newblock Gans trained by a two time-scale update rule converge to a local nash equilibrium, 2018.

\bibitem[Ho et~al.(2020)Ho, Jain, and Abbeel]{ho2020denoising}
Jonathan Ho, Ajay Jain, and Pieter Abbeel.
\newblock Denoising diffusion probabilistic models, 2020.

\bibitem[Ho et~al.(2022)Ho, Chan, Saharia, Whang, Gao, Gritsenko, Kingma, Poole, Norouzi, Fleet, et~al.]{ho2022imagen}
Jonathan Ho, William Chan, Chitwan Saharia, Jay Whang, Ruiqi Gao, Alexey Gritsenko, Diederik~P Kingma, Ben Poole, Mohammad Norouzi, David~J Fleet, et~al.
\newblock Imagen video: High definition video generation with diffusion models.
\newblock \emph{arXiv preprint arXiv:2210.02303}, 2022.

\bibitem[Hyv{\"a}rinen and Dayan(2005)]{hyvarinen2005estimation}
Aapo Hyv{\"a}rinen and Peter Dayan.
\newblock Estimation of non-normalized statistical models by score matching.
\newblock \emph{Journal of Machine Learning Research}, 6\penalty0 (4), 2005.

\bibitem[Iglesias et~al.(2023)Iglesias, Talavera, and Díaz-Álvarez]{Iglesias_2023}
Guillermo Iglesias, Edgar Talavera, and Alberto Díaz-Álvarez.
\newblock A survey on gans for computer vision: Recent research, analysis and taxonomy.
\newblock \emph{Computer Science Review}, 48:\penalty0 100553, 2023.

\bibitem[Iizuka et~al.(2017)Iizuka, Simo-Serra, and Ishikawa]{Iizuka}
Satoshi Iizuka, Edgar Simo-Serra, and Hiroshi Ishikawa.
\newblock Globally and locally consistent image completion.
\newblock \emph{ACM Trans. Graph.}, 36\penalty0 (4), 2017.

\bibitem[Isola et~al.(2018)Isola, Zhu, Zhou, and Efros]{isola2018imagetoimage}
Phillip Isola, Jun-Yan Zhu, Tinghui Zhou, and Alexei~A. Efros.
\newblock Image-to-image translation with conditional adversarial networks, 2018.

\bibitem[Kang et~al.(2023)Kang, Zhu, Zhang, Park, Shechtman, Paris, and Park]{kang2023scaling}
Minguk Kang, Jun-Yan Zhu, Richard Zhang, Jaesik Park, Eli Shechtman, Sylvain Paris, and Taesung Park.
\newblock Scaling up gans for text-to-image synthesis, 2023.

\bibitem[Karras et~al.(2019)Karras, Laine, and Aila]{karras2019stylebased}
Tero Karras, Samuli Laine, and Timo Aila.
\newblock A style-based generator architecture for generative adversarial networks, 2019.

\bibitem[Kingma and Welling(2019)]{Kingma_2019}
Diederik~P. Kingma and Max Welling.
\newblock An introduction to variational autoencoders.
\newblock \emph{Foundations and Trends® in Machine Learning}, 12\penalty0 (4):\penalty0 307–392, 2019.

\bibitem[Kingma and Welling(2022)]{kingma2022autoencoding}
Diederik~P Kingma and Max Welling.
\newblock Auto-encoding variational bayes, 2022.

\bibitem[Kobyzev et~al.(2021)Kobyzev, Prince, and Brubaker]{Kobyzev_2021}
Ivan Kobyzev, Simon~J.D. Prince, and Marcus~A. Brubaker.
\newblock Normalizing flows: An introduction and review of current methods.
\newblock \emph{IEEE Transactions on Pattern Analysis and Machine Intelligence}, 43\penalty0 (11):\penalty0 3964–3979, 2021.

\bibitem[Kudo and Richardson(2018)]{kudo2018sentencepiece}
Taku Kudo and John Richardson.
\newblock Sentencepiece: A simple and language independent subword tokenizer and detokenizer for neural text processing, 2018.

\bibitem[Kynkäänniemi et~al.(2019)Kynkäänniemi, Karras, Laine, Lehtinen, and Aila]{kynkäänniemi2019improved}
Tuomas Kynkäänniemi, Tero Karras, Samuli Laine, Jaakko Lehtinen, and Timo Aila.
\newblock Improved precision and recall metric for assessing generative models, 2019.

\bibitem[Larochelle and Murray(2011)]{pmlr-v15-larochelle11a}
Hugo Larochelle and Iain Murray.
\newblock The neural autoregressive distribution estimator.
\newblock In \emph{Proceedings of the Fourteenth International Conference on Artificial Intelligence and Statistics}, pages 29--37, Fort Lauderdale, FL, USA, 2011. PMLR.

\bibitem[Lecun et~al.(2006)Lecun, Chopra, and Hadsell]{yann}
Yann Lecun, Sumit Chopra, and Raia Hadsell.
\newblock \emph{A tutorial on energy-based learning}.
\newblock 2006.

\bibitem[Ledig et~al.(2017)Ledig, Theis, Huszar, Caballero, Cunningham, Acosta, Aitken, Tejani, Totz, Wang, and Shi]{ledig2017photorealistic}
Christian Ledig, Lucas Theis, Ferenc Huszar, Jose Caballero, Andrew Cunningham, Alejandro Acosta, Andrew Aitken, Alykhan Tejani, Johannes Totz, Zehan Wang, and Wenzhe Shi.
\newblock Photo-realistic single image super-resolution using a generative adversarial network, 2017.

\bibitem[Li et~al.(2019)Li, Qi, Lukasiewicz, and Torr]{li2019controllable}
Bowen Li, Xiaojuan Qi, Thomas Lukasiewicz, and Philip H.~S. Torr.
\newblock Controllable text-to-image generation, 2019.

\bibitem[Li et~al.(2022)Li, Yang, Chang, Chen, Feng, Xu, Li, and Chen]{LI202247}
Haoying Li, Yifan Yang, Meng Chang, Shiqi Chen, Huajun Feng, Zhihai Xu, Qi Li, and Yueting Chen.
\newblock Srdiff: Single image super-resolution with diffusion probabilistic models.
\newblock \emph{Neurocomputing}, 479:\penalty0 47--59, 2022.

\bibitem[Lugmayr et~al.(2022)Lugmayr, Danelljan, Romero, Yu, Timofte, and Van~Gool]{Lugmayr_2022_CVPR}
Andreas Lugmayr, Martin Danelljan, Andres Romero, Fisher Yu, Radu Timofte, and Luc Van~Gool.
\newblock Repaint: Inpainting using denoising diffusion probabilistic models.
\newblock In \emph{Proceedings of the IEEE/CVF Conference on Computer Vision and Pattern Recognition (CVPR)}, pages 11461--11471, 2022.

\bibitem[Luo and Hu(2021)]{luo2021diffusion}
Shitong Luo and Wei Hu.
\newblock Diffusion probabilistic models for 3d point cloud generation.
\newblock In \emph{Proceedings of the IEEE/CVF Conference on Computer Vision and Pattern Recognition}, pages 2837--2845, 2021.

\bibitem[Mansimov et~al.(2016)Mansimov, Parisotto, Ba, and Salakhutdinov]{mansimov2016generating}
Elman Mansimov, Emilio Parisotto, Jimmy~Lei Ba, and Ruslan Salakhutdinov.
\newblock Generating images from captions with attention, 2016.

\bibitem[Mirza and Osindero(2014)]{mirza2014conditional}
Mehdi Mirza and Simon Osindero.
\newblock Conditional generative adversarial nets, 2014.

\bibitem[Miyato et~al.(2018)Miyato, Kataoka, Koyama, and Yoshida]{miyato2018spectral}
Takeru Miyato, Toshiki Kataoka, Masanori Koyama, and Yuichi Yoshida.
\newblock Spectral normalization for generative adversarial networks, 2018.

\bibitem[Ngiam et~al.(2011)Ngiam, Chen, Koh, and Ng]{jiquan}
Jiquan Ngiam, Zhenghao Chen, Pang Koh, and Andrew Ng.
\newblock Learning deep energy models.
\newblock pages 1105--1112, 2011.

\bibitem[Nichol et~al.(2022{\natexlab{a}})Nichol, Dhariwal, Ramesh, Shyam, Mishkin, McGrew, Sutskever, and Chen]{glide}
Alex Nichol, Prafulla Dhariwal, Aditya Ramesh, Pranav Shyam, Pamela Mishkin, Bob McGrew, Ilya Sutskever, and Mark Chen.
\newblock Glide: Towards photorealistic image generation and editing with text-guided diffusion models, 2022{\natexlab{a}}.

\bibitem[Nichol et~al.(2022{\natexlab{b}})Nichol, Dhariwal, Ramesh, Shyam, Mishkin, McGrew, Sutskever, and Chen]{nichol2022glide}
Alex Nichol, Prafulla Dhariwal, Aditya Ramesh, Pranav Shyam, Pamela Mishkin, Bob McGrew, Ilya Sutskever, and Mark Chen.
\newblock Glide: Towards photorealistic image generation and editing with text-guided diffusion models, 2022{\natexlab{b}}.

\bibitem[Nowozin et~al.(2016)Nowozin, Cseke, and Tomioka]{nowozin2016fgan}
Sebastian Nowozin, Botond Cseke, and Ryota Tomioka.
\newblock f-gan: Training generative neural samplers using variational divergence minimization, 2016.

\bibitem[Papamakarios et~al.(2021)Papamakarios, Nalisnick, Rezende, Mohamed, and Lakshminarayanan]{papamakarios2021normalizing}
George Papamakarios, Eric Nalisnick, Danilo~Jimenez Rezende, Shakir Mohamed, and Balaji Lakshminarayanan.
\newblock Normalizing flows for probabilistic modeling and inference, 2021.

\bibitem[Pathak et~al.(2016)Pathak, Krahenbuhl, Donahue, Darrell, and Efros]{pathak2016context}
Deepak Pathak, Philipp Krahenbuhl, Jeff Donahue, Trevor Darrell, and Alexei~A Efros.
\newblock Context encoders: Feature learning by inpainting.
\newblock In \emph{Proceedings of the IEEE conference on computer vision and pattern recognition}, pages 2536--2544, 2016.

\bibitem[Radford et~al.(2016)Radford, Metz, and Chintala]{radford2016unsupervised}
Alec Radford, Luke Metz, and Soumith Chintala.
\newblock Unsupervised representation learning with deep convolutional generative adversarial networks, 2016.

\bibitem[Radford et~al.(2021)Radford, Kim, Hallacy, Ramesh, Goh, Agarwal, Sastry, Askell, Mishkin, Clark, Krueger, and Sutskever]{radford2021learning}
Alec Radford, Jong~Wook Kim, Chris Hallacy, Aditya Ramesh, Gabriel Goh, Sandhini Agarwal, Girish Sastry, Amanda Askell, Pamela Mishkin, Jack Clark, Gretchen Krueger, and Ilya Sutskever.
\newblock Learning transferable visual models from natural language supervision, 2021.

\bibitem[Ramesh et~al.(2021)Ramesh, Pavlov, Goh, Gray, Voss, Radford, Chen, and Sutskever]{ramesh2021zeroshot}
Aditya Ramesh, Mikhail Pavlov, Gabriel Goh, Scott Gray, Chelsea Voss, Alec Radford, Mark Chen, and Ilya Sutskever.
\newblock Zero-shot text-to-image generation, 2021.

\bibitem[Ramesh et~al.(2022{\natexlab{a}})Ramesh, Dhariwal, Nichol, Chu, and Chen]{dalle2}
Aditya Ramesh, Prafulla Dhariwal, Alex Nichol, Casey Chu, and Mark Chen.
\newblock Hierarchical text-conditional image generation with clip latents, 2022{\natexlab{a}}.

\bibitem[Ramesh et~al.(2022{\natexlab{b}})Ramesh, Dhariwal, Nichol, Chu, and Chen]{ramesh2022hierarchical}
Aditya Ramesh, Prafulla Dhariwal, Alex Nichol, Casey Chu, and Mark Chen.
\newblock Hierarchical text-conditional image generation with clip latents, 2022{\natexlab{b}}.

\bibitem[Reed et~al.(2016)Reed, Akata, Yan, Logeswaran, Schiele, and Lee]{reed2016generative}
Scott Reed, Zeynep Akata, Xinchen Yan, Lajanugen Logeswaran, Bernt Schiele, and Honglak Lee.
\newblock Generative adversarial text to image synthesis, 2016.

\bibitem[Ren et~al.(2019)Ren, Yu, Zhang, Li, Liu, and Li]{ren2019structureflow}
Yurui Ren, Xiaoming Yu, Ruonan Zhang, Thomas~H Li, Shan Liu, and Ge Li.
\newblock Structureflow: Image inpainting via structure-aware appearance flow.
\newblock In \emph{Proceedings of the IEEE/CVF international conference on computer vision}, pages 181--190, 2019.

\bibitem[Rezende and Mohamed(2016)]{rezende2016variational}
Danilo~Jimenez Rezende and Shakir Mohamed.
\newblock Variational inference with normalizing flows, 2016.

\bibitem[Rombach et~al.(2022)Rombach, Blattmann, Lorenz, Esser, and Ommer]{rombach2022highresolution}
Robin Rombach, Andreas Blattmann, Dominik Lorenz, Patrick Esser, and Björn Ommer.
\newblock High-resolution image synthesis with latent diffusion models, 2022.

\bibitem[Rudolph et~al.(2020)Rudolph, Wandt, and Rosenhahn]{rudolph2020differnet}
Marco Rudolph, Bastian Wandt, and Bodo Rosenhahn.
\newblock Same same but differnet: Semi-supervised defect detection with normalizing flows, 2020.

\bibitem[Saharia et~al.(2021)Saharia, Ho, Chan, Salimans, Fleet, and Norouzi]{saharia2021image}
Chitwan Saharia, Jonathan Ho, William Chan, Tim Salimans, David~J. Fleet, and Mohammad Norouzi.
\newblock Image super-resolution via iterative refinement, 2021.

\bibitem[Saharia et~al.(2022{\natexlab{a}})Saharia, Chan, Chang, Lee, Ho, Salimans, Fleet, and Norouzi]{saharia2022palette}
Chitwan Saharia, William Chan, Huiwen Chang, Chris~A. Lee, Jonathan Ho, Tim Salimans, David~J. Fleet, and Mohammad Norouzi.
\newblock Palette: Image-to-image diffusion models, 2022{\natexlab{a}}.

\bibitem[Saharia et~al.(2022{\natexlab{b}})Saharia, Chan, Saxena, Li, Whang, Denton, Ghasemipour, Ayan, Mahdavi, Lopes, Salimans, Ho, Fleet, and Norouzi]{saharia2022photorealistic}
Chitwan Saharia, William Chan, Saurabh Saxena, Lala Li, Jay Whang, Emily Denton, Seyed Kamyar~Seyed Ghasemipour, Burcu~Karagol Ayan, S.~Sara Mahdavi, Rapha~Gontijo Lopes, Tim Salimans, Jonathan Ho, David~J Fleet, and Mohammad Norouzi.
\newblock Photorealistic text-to-image diffusion models with deep language understanding, 2022{\natexlab{b}}.

\bibitem[Sajjadi et~al.(2018)Sajjadi, Bachem, Lucic, Bousquet, and Gelly]{sajjadi2018assessing}
Mehdi S.~M. Sajjadi, Olivier Bachem, Mario Lucic, Olivier Bousquet, and Sylvain Gelly.
\newblock Assessing generative models via precision and recall, 2018.

\bibitem[Salimans et~al.(2016)Salimans, Goodfellow, Zaremba, Cheung, Radford, and Chen]{salimans2016improved}
Tim Salimans, Ian Goodfellow, Wojciech Zaremba, Vicki Cheung, Alec Radford, and Xi Chen.
\newblock Improved techniques for training gans, 2016.

\bibitem[Salimans et~al.(2017)Salimans, Karpathy, Chen, and Kingma]{salimans2017pixelcnn}
Tim Salimans, Andrej Karpathy, Xi Chen, and Diederik~P. Kingma.
\newblock Pixelcnn++: Improving the pixelcnn with discretized logistic mixture likelihood and other modifications, 2017.

\bibitem[Sauer et~al.(2023)Sauer, Karras, Laine, Geiger, and Aila]{sauer2023stylegant}
Axel Sauer, Tero Karras, Samuli Laine, Andreas Geiger, and Timo Aila.
\newblock Stylegan-t: Unlocking the power of gans for fast large-scale text-to-image synthesis, 2023.

\bibitem[Schlegl et~al.(2017)Schlegl, Seeböck, Waldstein, Schmidt-Erfurth, and Langs]{schlegl2017unsupervised}
Thomas Schlegl, Philipp Seeböck, Sebastian~M. Waldstein, Ursula Schmidt-Erfurth, and Georg Langs.
\newblock Unsupervised anomaly detection with generative adversarial networks to guide marker discovery, 2017.

\bibitem[Schlegl et~al.(2019)Schlegl, Seeböck, Waldstein, Langs, and Schmidt-Erfurth]{SCHLEGL201930}
Thomas Schlegl, Philipp Seeböck, Sebastian~M. Waldstein, Georg Langs, and Ursula Schmidt-Erfurth.
\newblock f-anogan: Fast unsupervised anomaly detection with generative adversarial networks.
\newblock \emph{Medical Image Analysis}, 54:\penalty0 30--44, 2019.

\bibitem[Shang et~al.(2020)Shang, Dai, Zhu, Yang, and Guo]{shang2020perceptual}
Taizhang Shang, Qiuju Dai, Shengchen Zhu, Tong Yang, and Yandong Guo.
\newblock Perceptual extreme super resolution network with receptive field block, 2020.

\bibitem[Sohl-Dickstein et~al.(2015)Sohl-Dickstein, Weiss, Maheswaranathan, and Ganguli]{sohldickstein2015deep}
Jascha Sohl-Dickstein, Eric~A. Weiss, Niru Maheswaranathan, and Surya Ganguli.
\newblock Deep unsupervised learning using nonequilibrium thermodynamics, 2015.

\bibitem[Song and Ermon(2020{\natexlab{a}})]{song2020generative}
Yang Song and Stefano Ermon.
\newblock Generative modeling by estimating gradients of the data distribution, 2020{\natexlab{a}}.

\bibitem[Song and Ermon(2020{\natexlab{b}})]{song2020improved}
Yang Song and Stefano Ermon.
\newblock Improved techniques for training score-based generative models, 2020{\natexlab{b}}.

\bibitem[Song and Kingma(2021)]{song2021train}
Yang Song and Diederik~P. Kingma.
\newblock How to train your energy-based models, 2021.

\bibitem[Song et~al.(2019)Song, Garg, Shi, and Ermon]{song2019sliced}
Yang Song, Sahaj Garg, Jiaxin Shi, and Stefano Ermon.
\newblock Sliced score matching: A scalable approach to density and score estimation, 2019.

\bibitem[Song et~al.(2021)Song, Sohl-Dickstein, Kingma, Kumar, Ermon, and Poole]{song2021scorebased}
Yang Song, Jascha Sohl-Dickstein, Diederik~P. Kingma, Abhishek Kumar, Stefano Ermon, and Ben Poole.
\newblock Score-based generative modeling through stochastic differential equations, 2021.

\bibitem[Szegedy et~al.(2014)Szegedy, Liu, Jia, Sermanet, Reed, Anguelov, Erhan, Vanhoucke, and Rabinovich]{szegedy2014going}
Christian Szegedy, Wei Liu, Yangqing Jia, Pierre Sermanet, Scott Reed, Dragomir Anguelov, Dumitru Erhan, Vincent Vanhoucke, and Andrew Rabinovich.
\newblock Going deeper with convolutions, 2014.

\bibitem[Tumanyan et~al.(2023)Tumanyan, Geyer, Bagon, and Dekel]{tumanyan2023plug}
Narek Tumanyan, Michal Geyer, Shai Bagon, and Tali Dekel.
\newblock Plug-and-play diffusion features for text-driven image-to-image translation.
\newblock In \emph{Proceedings of the IEEE/CVF Conference on Computer Vision and Pattern Recognition}, pages 1921--1930, 2023.

\bibitem[Uria et~al.(2016)Uria, Côté, Gregor, Murray, and Larochelle]{uria2016neural}
Benigno Uria, Marc-Alexandre Côté, Karol Gregor, Iain Murray, and Hugo Larochelle.
\newblock Neural autoregressive distribution estimation, 2016.

\bibitem[van~den Oord et~al.(2016{\natexlab{a}})van~den Oord, Dieleman, Zen, Simonyan, Vinyals, Graves, Kalchbrenner, Senior, and Kavukcuoglu]{oord2016wavenet}
Aaron van~den Oord, Sander Dieleman, Heiga Zen, Karen Simonyan, Oriol Vinyals, Alex Graves, Nal Kalchbrenner, Andrew Senior, and Koray Kavukcuoglu.
\newblock Wavenet: A generative model for raw audio, 2016{\natexlab{a}}.

\bibitem[van~den Oord et~al.(2016{\natexlab{b}})van~den Oord, Kalchbrenner, and Kavukcuoglu]{oord2016pixel}
Aaron van~den Oord, Nal Kalchbrenner, and Koray Kavukcuoglu.
\newblock Pixel recurrent neural networks, 2016{\natexlab{b}}.

\bibitem[van~den Oord et~al.(2016{\natexlab{c}})van~den Oord, Kalchbrenner, Vinyals, Espeholt, Graves, and Kavukcuoglu]{oord2016conditional}
Aaron van~den Oord, Nal Kalchbrenner, Oriol Vinyals, Lasse Espeholt, Alex Graves, and Koray Kavukcuoglu.
\newblock Conditional image generation with pixelcnn decoders, 2016{\natexlab{c}}.

\bibitem[van~den Oord et~al.(2018)van~den Oord, Vinyals, and Kavukcuoglu]{oord2018neural}
Aaron van~den Oord, Oriol Vinyals, and Koray Kavukcuoglu.
\newblock Neural discrete representation learning, 2018.

\bibitem[Vaswani et~al.(2023)Vaswani, Shazeer, Parmar, Uszkoreit, Jones, Gomez, Kaiser, and Polosukhin]{vaswani2023attention}
Ashish Vaswani, Noam Shazeer, Niki Parmar, Jakob Uszkoreit, Llion Jones, Aidan~N. Gomez, Lukasz Kaiser, and Illia Polosukhin.
\newblock Attention is all you need, 2023.

\bibitem[Wang et~al.(2018)Wang, Yu, Wu, Gu, Liu, Dong, Loy, Qiao, and Tang]{wang2018esrgan}
Xintao Wang, Ke Yu, Shixiang Wu, Jinjin Gu, Yihao Liu, Chao Dong, Chen~Change Loy, Yu Qiao, and Xiaoou Tang.
\newblock Esrgan: Enhanced super-resolution generative adversarial networks, 2018.

\bibitem[Welling and Teh(2011)]{welling2011bayesian}
Max Welling and Yee~W Teh.
\newblock Bayesian learning via stochastic gradient langevin dynamics.
\newblock In \emph{Proceedings of the 28th international conference on machine learning (ICML-11)}, pages 681--688, 2011.

\bibitem[Wyatt et~al.(2022)Wyatt, Leach, Schmon, and Willcocks]{9857019}
Julian Wyatt, Adam Leach, Sebastian~M. Schmon, and Chris~G. Willcocks.
\newblock Anoddpm: Anomaly detection with denoising diffusion probabilistic models using simplex noise.
\newblock In \emph{2022 IEEE/CVF Conference on Computer Vision and Pattern Recognition Workshops (CVPRW)}, pages 649--655, 2022.

\bibitem[Xiao et~al.(2024)Xiao, Yuan, Jiang, He, Jin, and Zhang]{10353979}
Yi Xiao, Qiangqiang Yuan, Kui Jiang, Jiang He, Xianyu Jin, and Liangpei Zhang.
\newblock Ediffsr: An efficient diffusion probabilistic model for remote sensing image super-resolution.
\newblock \emph{IEEE Transactions on Geoscience and Remote Sensing}, 62:\penalty0 1--14, 2024.

\bibitem[Yang et~al.(2023)Yang, Zhang, Song, Hong, Xu, Zhao, Zhang, Cui, and Yang]{yang2023diffusion}
Ling Yang, Zhilong Zhang, Yang Song, Shenda Hong, Runsheng Xu, Yue Zhao, Wentao Zhang, Bin Cui, and Ming-Hsuan Yang.
\newblock Diffusion models: A comprehensive survey of methods and applications, 2023.

\bibitem[Yi et~al.(2019)Yi, Walia, and Babyn]{Yi_2019}
Xin Yi, Ekta Walia, and Paul Babyn.
\newblock Generative adversarial network in medical imaging: A review.
\newblock \emph{Medical Image Analysis}, 58:\penalty0 101552, 2019.

\bibitem[Yu et~al.(2018)Yu, Lin, Yang, Shen, Lu, and Huang]{Yu_2018_CVPR}
Jiahui Yu, Zhe Lin, Jimei Yang, Xiaohui Shen, Xin Lu, and Thomas~S. Huang.
\newblock Generative image inpainting with contextual attention.
\newblock In \emph{Proceedings of the IEEE Conference on Computer Vision and Pattern Recognition (CVPR)}, 2018.

\bibitem[Yu et~al.(2021)Yu, Zheng, Wang, Li, Wu, Zhao, and Wu]{yu2021fastflow}
Jiawei Yu, Ye Zheng, Xiang Wang, Wei Li, Yushuang Wu, Rui Zhao, and Liwei Wu.
\newblock Fastflow: Unsupervised anomaly detection and localization via 2d normalizing flows, 2021.

\bibitem[Yu et~al.(2022{\natexlab{a}})Yu, Li, Koh, Zhang, Pang, Qin, Ku, Xu, Baldridge, and Wu]{yu2022vectorquantized}
Jiahui Yu, Xin Li, Jing~Yu Koh, Han Zhang, Ruoming Pang, James Qin, Alexander Ku, Yuanzhong Xu, Jason Baldridge, and Yonghui Wu.
\newblock Vector-quantized image modeling with improved vqgan, 2022{\natexlab{a}}.

\bibitem[Yu et~al.(2022{\natexlab{b}})Yu, Xu, Koh, Luong, Baid, Wang, Vasudevan, Ku, Yang, Ayan, Hutchinson, Han, Parekh, Li, Zhang, Baldridge, and Wu]{parti}
Jiahui Yu, Yuanzhong Xu, Jing~Yu Koh, Thang Luong, Gunjan Baid, Zirui Wang, Vijay Vasudevan, Alexander Ku, Yinfei Yang, Burcu~Karagol Ayan, Ben Hutchinson, Wei Han, Zarana Parekh, Xin Li, Han Zhang, Jason Baldridge, and Yonghui Wu.
\newblock Scaling autoregressive models for content-rich text-to-image generation, 2022{\natexlab{b}}.

\bibitem[Yu et~al.(2022{\natexlab{c}})Yu, Xu, Koh, Luong, Baid, Wang, Vasudevan, Ku, Yang, Ayan, Hutchinson, Han, Parekh, Li, Zhang, Baldridge, and Wu]{yu2022scaling}
Jiahui Yu, Yuanzhong Xu, Jing~Yu Koh, Thang Luong, Gunjan Baid, Zirui Wang, Vijay Vasudevan, Alexander Ku, Yinfei Yang, Burcu~Karagol Ayan, Ben Hutchinson, Wei Han, Zarana Parekh, Xin Li, Han Zhang, Jason Baldridge, and Yonghui Wu.
\newblock Scaling autoregressive models for content-rich text-to-image generation, 2022{\natexlab{c}}.

\bibitem[Zhai et~al.(2016)Zhai, Cheng, Lu, and Zhang]{zhai2016deep}
Shuangfei Zhai, Yu Cheng, Weining Lu, and Zhongfei Zhang.
\newblock Deep structured energy based models for anomaly detection.
\newblock In \emph{International conference on machine learning}, pages 1100--1109. PMLR, 2016.

\bibitem[Zhang et~al.(2023{\natexlab{a}})Zhang, Zhang, Zhang, and Kweon]{zhang2023texttoimage}
Chenshuang Zhang, Chaoning Zhang, Mengchun Zhang, and In~So Kweon.
\newblock Text-to-image diffusion models in generative ai: A survey, 2023{\natexlab{a}}.

\bibitem[Zhang et~al.(2023{\natexlab{b}})Zhang, Huang, Liu, Wang, and Jin]{zhang2023swinfir}
Dafeng Zhang, Feiyu Huang, Shizhuo Liu, Xiaobing Wang, and Zhezhu Jin.
\newblock Swinfir: Revisiting the swinir with fast fourier convolution and improved training for image super-resolution, 2023{\natexlab{b}}.

\bibitem[Zhang et~al.(2017)Zhang, Xu, Li, Zhang, Wang, Huang, and Metaxas]{zhang2017stackgan}
Han Zhang, Tao Xu, Hongsheng Li, Shaoting Zhang, Xiaogang Wang, Xiaolei Huang, and Dimitris Metaxas.
\newblock Stackgan: Text to photo-realistic image synthesis with stacked generative adversarial networks, 2017.

\bibitem[Zhang et~al.(2019)Zhang, Goodfellow, Metaxas, and Odena]{zhang2019selfattention}
Han Zhang, Ian Goodfellow, Dimitris Metaxas, and Augustus Odena.
\newblock Self-attention generative adversarial networks, 2019.

\bibitem[Zhang et~al.(2024)Zhang, Li, Zhou, Zhao, and Gu]{zhang2024transcending}
Leheng Zhang, Yawei Li, Xingyu Zhou, Xiaorui Zhao, and Shuhang Gu.
\newblock Transcending the limit of local window: Advanced super-resolution transformer with adaptive token dictionary, 2024.

\bibitem[Zhao et~al.(2017)Zhao, Mathieu, and LeCun]{zhao2017energybased}
Junbo Zhao, Michael Mathieu, and Yann LeCun.
\newblock Energy-based generative adversarial network, 2017.

\bibitem[Zhao et~al.(2021)Zhao, Cui, Sheng, Dong, Liang, Chang, and Xu]{zhao2021large}
Shengyu Zhao, Jonathan Cui, Yilun Sheng, Yue Dong, Xiao Liang, Eric~I Chang, and Yan Xu.
\newblock Large scale image completion via co-modulated generative adversarial networks, 2021.

\end{thebibliography}
